# Development of an Unpaired Deep Neural Network for Synthesizing X-ray Fluoroscopic Images from Digitally Reconstructed Tomography in Image Guided Radiotherapy

Chisako Hayashi[1,2], BS, Shinichiro Mori[1], PhD, Yasukuni Mori[3], Lim Taehyeung[1,2], BS, Hiroki Suyari[3], PhD, Hitoshi Ishikawa[4], MD-PhD

[1]Institute for Quantum Medical Science, National Institutes for Quantum Science and Technology, Inage-ku, Chiba 263-8555, Japan

[2]Graduate School of Science and Engineering, Chiba University, 263-8522, Japan

[3]Graduate School of Engineering, Chiba University, 263-8522, Japan

[4]QST hospital, National Institutes for Quantum Science and Technology, Inage-ku, Chiba 263-8555, Japan

Manuscript type: Full paper
Running title: Image synthetic thoracic X-ray image
Conflict of Interest Statement: No

Author for correspondence: Shinichiro Mori, PhD
Institute for Quantum Medical Science, National Institutes for Quantum Science and Technology, Inage-ku, Chiba 263-8555, Japan
Office: 81-43-251-2111, Fax: 81-43-284-0198
e-mail: mori.shinichiro@qst.go.jp

**ABSTRACT**

Purpose

The purpose of this study was to develop and evaluate a deep neural network (DNN) capable of generating flat-panel detector (FPD) images from digitally reconstructed radiography (DRR) images in lung cancer treatment, with the aim of improving clinical workflows in image-guided radiotherapy.

Methods

A modified CycleGAN architecture was trained on paired DRR–FPD image data obtained from patients with lung tumors. The training dataset consisted of over 400 DRR–FPD image pairs, and the final model was evaluated on an independent set of 100 FPD images. Mean absolute error (MAE), peak signal-to-noise ratio (PSNR), structural similarity index measure (SSIM), and Kernel Inception Distance (KID) were used to quantify the similarity between synthetic and ground-truth FPD images. Computation time for generating synthetic images was also measured.

Results

Despite some positional mismatches in the DRR–FPD pairs, the synthetic FPD images closely resembled the ground-truth FPD images. The proposed DNN achieved notable improvements over both input DRR images and a U-Net–based method in terms of MAE, PSNR, SSIM, and KID. The average image generation time was on the order of milliseconds per image, indicating its potential for real-time application. Qualitative evaluations showed that the DNN successfully reproduced image noise patterns akin to real FPD images, reducing the need for manual noise adjustments.

Conclusions

The proposed DNN effectively converted DRR images into realistic FPD images for thoracic cases, offering a fast and practical method that could streamline patient setup verification and enhance overall clinical workflow. Future work should validate the model across different imaging systems and address remaining challenges in marker visualization, thereby fostering broader clinical adoption.

**Keywords:** Deep neural network, Image quality, Intrafractional motion, Radiotherapy

# I. INTRODUCTION

Recent advancements in deep neural network (DNN) technologies have significantly accelerated progress in medical image processing, often surpassing the capabilities of traditional image processing methods (1-3). These developments have led to notable improvements in various areas of radiation therapy, including image-guided radiation therapy (IGRT), treatment planning, auto-segmentation (4, 5), automated treatment planning (6), synthetic image generation (7), deformable image registration (8), and real-time tumor tracking techniques (9, 10). By enhancing automation and accuracy, these innovations aim to optimize both treatment precision and workflow efficiency in clinical practice. A potential approach to further improve treatment accuracy is to standardize both images under the same imaging modality, a process commonly referred to as image synthesis. Image synthesis techniques can be categorized into two types: intra-modality and inter-modality synthesis. Intra-modality synthesis involves transforming an image into another form within the same modality, such as MRI to CT (11, 12), PET to CT (13, 14)、CT to MRI (15, 16).

Previously, our team developed a DNN designed to generate X-ray flat panel detector (FPD) images from digitally reconstructed radiography (DRR) images for the prostate and head-and-neck regions (17). However, residual interfractional organ shifts persisted due to differences in image acquisition timing. To mitigate these variations, a possible approach is to create mimic FPD images using Monte Carlo simulation-based DRR calculations, which could reduce anatomical inconsistencies. However, since the quality of mimic FPD images may not fully match original FPD images, their effectiveness in training a DNN for high-quality synthetic FPD generation remains uncertain. Using synthetic FPD images could potentially accelerate patient positioning verification, as aligning FPD images with reference DRR images remains time-consuming, often requiring 10–20 seconds for 2D/3D auto-registration calculations but up to 2–5 minutes for final positional verification (18). This delay arises from the challenges of visually comparing images from different modalities with varying image qualities. However, this technique may be most effective in anatomical regions with minimal interfractional variations.

Our hospital utilizes real-time tumor tracking without implanted fiducial markers for thoracoabdominal treatments (19). One of our markerless tracking systems employs a multi-template matching algorithm (20), which uses FPD images acquired over several respiratory cycles after patient setup. However, these FPD images do not directly contain tumor position information, making manual tumor contouring labor-intensive due to the large number of images. To address this, tumor contours on FPD images can be automatically generated using 4D-DRR images, as tumor contours are typically defined on the 4DCT dataset during treatment planning, often with deformable image registration. Since image registration errors cannot be completely eliminated, medical staff can manually adjust tumor positions on the template images when necessary to improve tracking accuracy. A possible solution to this challenge is the use of pre-generated 4D-DRR images with tumor contours as multi-template images. This approach could significantly reduce the need for manual tumor delineation, streamlining the process.

With this goal in mind, we developed a DNN based on a modified CycleGAN (21) to generate FPD images from DRR images, even when interfractional anatomical variations were present. The quality of the synthetic FPD images was compared with both original FPD and DRR images from lung cases to evaluate the performance of the proposed method.

## II. MATERIALS AND METHODS

### II.A. Patients and Image Acquisition

At our center, 107 patients with lung tumors received carbon-ion beam treatment. The study was conducted with the approval of the Institutional Review Board (N20-044) and performed in accordance with the Declaration of Helsinki. All the patients provided informed consent for use the data from their medical records. During image acquisition, all patients were positioned on the treatment table with immobilization devices (urethane resin cushion [Moldcare, Alcare, Tokyo, Japan]) and low-temperature thermoplastic shells (Shell Fitter, Kuraray Co., Ltd., Osaka, Japan).

### II.A.1 Planning CT image

Treatment planning CT scans were performed during free-breathing using a 320-detector CT scanner (Aquilion One Vision®, Canon Medical Systems, Otawara, Japan). To accommodate the orthogonal beam arrangements (0° and 90°) used in our facility, the treatment couch was rotated along its longitudinal axis to extend the beam angle options or adjusted to shift the patient between prone and supine positions. As a result, 107 patients underwent a single 4DCT scan, while 48 patients required two 4DCT scans to account for different couch orientations or patient positions.

The 4DCT imaging was conducted with a tube voltage of 120 kV and a slice collimation of either $270 \times 0.5$ mm or $320 \times 0.5$ mm in volumetric cine mode. Continuous scanning was performed to cover the full extent of the lung region, and tube current modulation was used to maintain clinically acceptable image quality. The reconstructed parameters included a field of view (FOV) of 500 mm and a slice thickness of 2.0 mm. Each 4DCT dataset was divided into 10 respiratory phases (T00 representing peak inhalation and T50 representing peak exhalation).

*II.A.2 Fluoroscopic images*

Digital fluoroscopic images for lung treatments were captured using imaging systems installed in the treatment room (22). The X-ray tube was positioned 239 cm from the flat-panel detector (FPD) and 169 cm from the room isocenter. The DRR image matrix size and pixel size were set to 768 × 768 pixels and 388 × 388 μm, respectively.

Before initiating treatment, 2D–3D image registration was conducted using a pair of FPD images and the planning CT data to verify patient setup accuracy (23). This process involved aligning anatomical structures on the FPD images with those on the DRR images. During treatment, real-time acquisition of FPD images was performed to monitor the tumor position (19). This involved matching anatomical structures on the FPD images to their counterparts on the DRR images (24). To accommodate the lung treatment protocol, which requires the use of three or more beam angles, the treatment couch was rotated along its longitudinal axis ($\phi$, defined as the International Electrotechnical Commission [IEC] tabletop rolling angle). This rotation expanded the range of beam angles to cover −20° to +20°.

## II.C Network architecture

The original CycleGAN is optimized using adversarial loss, a fundamental component of GAN, along with a cycle consistency constraint. While CycleGAN is typically used for unsupervised training with unpaired data, our DNN was a modified CycleGAN trained with paired image data to generate an FPD image from a DRR image (Figure 1). The basic structure of a GAN consists of a confrontation between a Generator and a Discriminator. The Discriminator is trained to accurately distinguish whether the input is a real FPD image or one generated by the Generator. In this framework, the Generator's objective is to deceive the Discriminator, while the Discriminator's goal is to resist being deceived by the Generator. This competitive interaction enables both networks to improve, resulting in more precise image transformations.

*II.C.1 Image generator network*

Our network structure for the image generator, shown in Figure 2a, involves both encoding and decoding procedures.

The encoder block extracts features representing the input data while reducing spatial dimensions through a combination of the following layers: a convolutional (Conv) layer, a rectified linear unit (ReLU) activation layer, and an instance normalization (IN) layer (25). A reflect padding (RP) layer was incorporated at the initial stage to extend the image boundaries by mirroring the existing pixels, helping preserve image features and prevent edge artifacts during convolution operations. The network consists of four sequential groups, each composed of Conv + IN + ReLU layers. The convolutional kernel size was set to 4×4 pixels for the first layer and 3×3 pixels for the remaining layers. The number of output channels for each Conv layer was 64, 128, 128, and 256, respectively. The residual block consists of an RP + Conv + IN + ReLU layer sequence, followed by another Conv and IN layer. A skip connection was applied before the first RP layer and after the second Conv layer using an addition operation. The residual blocks were repeated nine times throughout the network.

Regarding the decoder block, two sets of deconvolution (Deconv) + IN + ReLU layers were added, with a deconvolutional kernel size of 3×3 pixels. The number of output channels was 128 for the first Deconv and 64 for the second. Finally, an RP + Conv + Sigmoid layer sequence was added. The convolutional kernel size and the number of output channels were set to 7×7 pixels and 1, respectively.

*II.C.2 Discriminator network*

The discriminator network was designed as a classifier to determine whether the input image was real or synthetic. As illustrated in Figure 2b, the network consists of four sequential groups, each composed of a convolutional layer, a Leaky ReLU activation layer, and an instance normalization layer with the exception of the first group, which lacks normalization, this is because sample oscillation and model instability could be reduced (26). The convolutional kernel size and stride were set to 4×4 pixels and 2×2 pixels, respectively, for the first three groups. The number of output channels was progressively increased, with the fourth convolutional layer outputting 512 channels— eight times the number of channels in the first convolutional layer (64 channels). Finally, a convolutional layer with a single output channel was added as the last layer. This multi-stage design enables the network to capture increasingly complex features as the data passes through deeper layers.

## II.D Network training
### *II.D.1 Training data*
### Projecting DRR image

A pair of DRR images was generated by projecting CT data, converted to X-ray attenuation coefficients, along the X-ray imaging beam path using our custom software (27). The projection was calculated as:

$$q(x, y) = \sum_{k=1}^{n} \Delta L \cdot \mu_k \qquad (1)$$

where $q(x, y)$ is the projection ray sum point on the DRR image position $(x, y)$ and $\Delta L$ is the calculation grid size (= 1 mm in this study).

To ensure the tumor was centered in the DRR image, the CT image was shifted accordingly. In cases where the edges of the DRR image did not fully encompass the CT image due to a limited number of slices, additional CT slices were appended before the first slice and after the last slice to extend the CT image region, improving DRR image quality. The DRR image matrix size and pixel size were set to $768 \times 768$ pixels and $388 \times 388$ μm, respectively, matching the dimensions of the FPD images. DRR computations were implemented using commercial software (Compute Unified Device Architecture [CUDA] version 10.1) with Microsoft Visual Studio 2013 (Microsoft Corp, Redmond, WA, USA) in a Windows 10 environment. The computations were performed on an NVIDIA Quadro A6000 GPU (NVIDIA Corporation, Santa Clara, CA, USA), featuring 10,752 CUDA cores and 48 GB of memory, enabling a processing speed exceeding 38.7 Tflops for single-precision calculations (28).

### Image preprocessing

A total of 404 image pairs (DRR and FPD images) were randomly selected for the DNN training process, respectively. To eliminate pixels with large errors, a pair of FPD and DRR images with a resolution of $768 \times 768$ pixels was cropped by removing 20 pixels from each edge—top, bottom, left, and right—resulting in a resolution of $728 \times 728$ pixels. Then, all FPD and DRR images were scaled to a final resolution of $384 \times 384$ pixels using bilinear interpolation. Additionally, the 304 image pairs were further subdivided into 20 subimages ($144 \times 144$ pixels each) per image by changing their position to ensure that the proportion of air in each subimage was less than 40%. Then, these subimages were augmented online by cropping them to $124 \times 124$ pixels with a random shift of $\pm 20$ pixels and applying random left-right or up-down flipping. The pixel values of all FPD and DRR images were normalized to a range of 0 to 1. Finally, a dataset of 6080 subimage pairs (from 79 cases) for training and 100 image pairs (from 28 cases) for evaluation was created.

### II.D.2 Parameter optimization

The DNN parameters were adjusted to predict FPD images from DRR images using the following procedure. The model was trained for 550 epochs with a batch size of 16, employing the Adamax optimizer (29) to minimize the loss function. The learning rate, beta1, beta2, and epsilon were initialized at $2 \times 10^{-4}$, 0.5, 0.999, and $10^{-7}$, respectively. Weight decay was not used. To refine the learning process, the learning rate was reduced to one-tenth after the 500 epoch. Although no formal early stopping criteria were established, training and validation loss curves were consistently monitored. Optimization was halted when the curves plateaued, or signs of overfitting emerged. The deep learning framework TensorFlow 2.12 was employed in a Windows 10, 64-bit environment, utilizing a single NVIDIA Quadro A6000 GPU.

Interfractional anatomical variations between the FPD and DRR images may cause discrepancies, preventing perfect alignment between the two modalities. To address this issue, four loss functions were applied to the image generator: adversarial loss ($L_{Adv}$), cycle-consistency loss ($L_{Cy}$), identity loss($L_{Id}$), and style loss ($L_{Sty}$). For the discriminator, a single adversarial loss was calculated, as described below:

$$\mathcal{L}_G = \arg \quad \min \left( \mathcal{L}_{Adv} + \lambda_1 \mathcal{L}_{Cyc} + \lambda_2 \mathcal{L}_{Id} + \lambda_3 \mathcal{L}_{Sty} \right).$$

$$\mathcal{L}_D = \arg \quad \max \mathcal{L}_{Adv}.$$

where $\lambda$ is weight factor. In this study, we set $\lambda_1$, $\lambda_2$ and $\lambda_3$ were 5.0, 5.0, and $2 \times 10^{-5}$, respectively.

### Adversarial loss

Adversarial loss was defined as

$$\mathcal{L}_{Adv}(G_{DRR-FPD}, D_{FPD}) = \mathbb{E}_{FPD}\big[log\big(D_{FPD}(I_{FPD})\big)\big] + \mathbb{E}_{DRR}\left[log\left(1 - D_{FPD}\big(G_{DRR-FPD}(I_{DRR})\big)\right)\right]$$

$$\mathcal{L}_{Adv}(G_{FPD-DRR}, D_{DRR}) = \mathbb{E}_{DRR}\big[log\big(D_{DRR}(I_{DRR})\big)\big] + \mathbb{E}_{FPD}\left[log\left(1 - D_{DRR}\big(G_{FPD-DRR}(I_{FPD})\big)\right)\right]$$

where $I_{FPD}$ and $I_{DRR}$ represent the FPD and DRR images, respectively. $G_{DRR-FPD}$ and $G_{FPD-DRR}$ refer to the image generators for generating $I_{FPD}$ from $I_{DRR}$ and $I_{DRR}$ from $I_{FPD}$, respectively.

### Cycle-consistency loss

Cycle-consistency loss is a loss function that ensure an image transformed into another domain returns to its original domain through inverse transformation, aiming to achieve reversibility in image conversion. The model is trained so that the image generated by $G_{DRR-FPD}$, when transformed back by $G_{FPD-DRR}$ (i.e., $G_{FPD-DRR}$ ($G_{DRR-FPD}$ ($I_{DRR}$))), closely resembles the original DRR image. It is expressed as follows;

$$\mathcal{L}_{Cyc}(G_{DRR-FPD}, G_{FPD-DRR})$$

$$= \mathbb{E}_{FPD}\left[\left\|G_{DRR-FPD}\big(G_{FPD-DRR}(I_{FPD})\big) - I_{FPD}\right\|_1\right]$$

$$+ \mathbb{E}_{DRR}\left[\left\|G_{FPD-DRR}\big(G_{DRR-FPD}(I_{DRR})\big) - I_{DRR}\right\|_1\right]$$

where $\|\cdot\|_1$ denotes L1 norm.

### Identity loss

Identity loss ($L_{Id}$) was introduced to achieve identity mapping within the same domain. The identity loss between $G_{DRR-FPD}(I_{DRR})$ and $I_{FPD}$, denoted as $L_{Id}$ ($G_{DRR-FPD}$), and the identity loss between $G_{FPD-DRR}(I_{FPD})$ and $I_{DRR}$, denoted as $L_{Id}$ ($G_{FPD-DRR}$) were defined as

$$\mathcal{L}_{Id}(G_{DRR-FPD}) = \mathbb{E}_{FPD}[\|G_{DRR-FPD}(I_{FPD}) - I_{FPD}\|_1]$$

$$\mathcal{L}_{Id}(G_{FPD-DRR}) = \mathbb{E}_{DRR}[\|G_{FPD-DRR}(I_{DRR}) - I_{DRR}\|_1]$$

### Style Transfer

In this study, a feature-level loss was introduced for style transfer in domain conversion. The style loss used for style transfer evaluates the degree of style matching between the output image $G_{DRR-FPD}(I_{DRR})$ and the ground-truth image $I_{FPD}$. By performing a comparison at the feature level, the goal is to promote style (i.e., domain) consistency without being affected by positional errors between the images.

Style loss was assessed using a pre-trained VGG19 model (30) as a feature extractor. It calculated the sum of six feature maps (from layers 1, 2, 5, 10, 15, and 20) extracted from the respective convolutional layers. The model was trained by computing the Gram matrices of these feature maps and minimizing the mean squared error (MSE) between them. The use of Gram matrices enables the model to capture pixel-wise correlations within the intermediate layers of VGG19, allowing the extraction of more spatially invariant features across a broader region.

The style loss between $G_{DRR-FPD}(I_{DRR})$ and $I_{FPD}$, denoted as $L_{Sty}$ ($G_{DRR-FPD}$), and the style loss between $G_{FPD-DRR}(I_{FPD})$ and $I_{DRR}$, denoted as $L_{Sty}$ ($G_{FPD-DRR}$) were defined as

$$\mathcal{L}_{Sty}(G_{DRR-FPD}) = E_{FPD}\left[\sum_i^{layers} \left\|Gram\left(V_i\big(G_{DRR-FPD}(I_{DRR})\big)\right) - Gram\big(V_i(I_{FPD})\big)\right\|_2\right].$$

$$\mathcal{L}_{Sty}(G_{FPD-DRR}) = E_{DRR}\left[\sum_i^{layers} \left\|Gram\left(V_i\big(G_{FPD-DRR}(I_{FPD})\big)\right) - Gram\big(V_i(I_{DRR})\big)\right\|_2\right].$$

$Gram(\cdot)$ was Gram matrices

## II.E Evaluations

We evaluated the quality of the synthetic FPD images using 100 ground-truth FPD images from 28 cases. These image datasets were independent of the training data. The synthetic FPD images were compared with the ground-truth FPD images using mean absolute error (MAE), peak signal-to-noise ratio (PSNR), and structural similarity index measure (SSIM) (31). These metrics are commonly used to assess the similarity between two images. Additionally, Kernel Inception Distance (KID) (32) was adopted as a metric to evaluate the distance between image distributions. Furthermore, we compared the image quality of the synthetic FPD images with that of the DRR images to further assess the performance of our DNN.

The computation time required for image prediction (excluding the model file import process) was also measured.

In this study, these evaluations were conducted after quantizing the proposed model.

## III. RESULTS

For case no. 4, the image quality of the synthetic FPD generated by our DNN (Figure 3c) from the DRR image (Figure 3a) was visually much closer to the ground-truth FPD (Figure 3b) compared to the synthetic FPD generated by U-Net (Figure 3d). The fiducial marker was also visualized clearly in the synthetic FPD image with our DNN compared to the U-Net (marked as yellow arrows in Figure 3). However, in regions where anatomical structures overlap and appear as small dark points, the DNN mistakenly identified them as fiducial markers (marked with green arrows in Figure 3c). The edge of the irradiation port cover was visible on the ground-truth FPD image (indicated by the blue arrow in Figure 3b). However, since this edge was not visible in the corresponding position on the DRR image, it was also not visualized in the synthetic FPD image. This was also the case for the patient call cable (indicated by the maganda arrow in Figure 3b). The quality of the input DRR image compared to the ground-truth FPD image was evaluated using the following metrics: MAE = 0.39, PSNR = 7.94 dB, and SSIM = 0.20. Our DNN improved these metrics to MAE = 0.04, PSNR = 26.85 dB, and SSIM = 0.79, outperforming U-Net (MAE = 0.36, PSNR = 8.86 dB, and SSIM = 0.20).

Since the tumor location for case no. 6 was in the lower lung lobe, most of the image region included the abdominal area rather than the lung (Figure 4). The image contrast in this region was lower, with greater image noise in the ground-truth FPD image (Figure 4b) compared to case no. 1. Despite this, the quality of the synthetic FPD image generated by our DNN (Figure 4c) was visually much closer to that of the ground-truth FPD image than the input DRR image (Figure 4a). The evaluation metrics, MAE and PSNR, for the synthetic FPD image with our DNN (0.08, 20.98) were improved compared to the DRR image (0.16, 14.52) and outperformed the synthetic FPD image with U-Net (0.15, 15.73). Although the SSIM value (0.39) was lower than that of the DRR image (0.44) and the FPD image generated by U-Net (0.41), this is likely because the synthetic FPD image generated by our DNN acquired noise components characteristic of the ground-truth FPD image.

Since the rib structure was visualized more clearly in the input DRR image than in the ground-truth FPD image, it also appeared clearer in the synthetic FPD image than in the ground-truth FPD image. Similar to case no. 4, the patient call cable was visible in the ground-truth FPD image but not in the synthetic FPD image, as it was absent from the input DRR image (indicated by the maganda arrow in Figure 4c). Additionally, bowel gas was visible in the synthetic FPD image because it was present in the input DRR image but absent from the ground-truth FPD image.

The results for image quality, averaged across all cases, are summarized in Figure 5 and Table 1. The interquartile range (IQR), representing the 25th to 75th percentiles, was improved for the synthetic FPD images generated by our DNN compared to both the DRR images and the synthetic FPD images generated by U-Net across all image quality metrics. MAE for the synthetic FPD image was reduced ($0.06 \pm 0.03$) compared to the input DRR image ($0.32 \pm 0.09$) (Figure 5a). The synthetic FPD image with our DNN also demonstrated a higher PSNR value ($24.47 \pm 5.09$ dB) than that with U-Net ($12.62 \pm 3.26$ dB) and the DRR image ($9.60 \pm 2.52$ dB) (Figure 5b). Similarly, the SSIM value ($0.32 \pm 0.16$) for the synthetic FPD images generated by our DNN was higher than the synthetic FPD images generated by U-Net ($0.35 \pm 0.16$) and the input DRR images ($0.32 \pm 0.16$) (Figure 5c). Additionally, the evaluation value based on KID was $0.7 \times 10^{-2}$, indicating that the data distribution of the synthetic FPD images generated by our DNN was closer to that of the real FPD images compared to the synthetic FPD images generated by U-Net ($1.7 \times 10^{-2}$) and the input DRR images ($5.1 \times 10^{-2}$). The average computation time was $12.9 \pm 5.2$ msec and $139.6 \pm 5.5$ msec for the synthetic FPD image with our DNN and U-Net, respectively.

## IV. DISCUSSION

We developed a DNN to generate synthetic FPD images from DRR images and evaluated the image quality by comparing the synthetic FPD images with the original FPD images using thoracic image data. Despite the presence of positional errors in the training data for both DRR and FPD images, the synthetic FPD images closely matched the quality of the original FPD images. The average computation time for image prediction was approximately 12.9 msec per image.

## IV.A. Image quality

In our previous study, the predicted FPD images did not fully capture the effects of scattered radiation. To address this, image noise was manually added to the predicted FPD images to better approximate real conditions (17). However, in this study, the DNN we developed was capable of generating image noise patterns more closely resembling those observed in real FPD images, reducing the need for manual noise adjustments and improving the overall realism of the synthetic images.

Our image quality results were lower compared to those reported in other studies(33, 34) using medical images such as CT and MRI. This difference can be attributed to the use of X-ray FPD images, where image quality is influenced by factors such as imaging conditions and system characteristics. Additionally, the FPD images in this study included elements do not present in the DRR images, such as bowel gas, the irradiation port edge, the patient call cable, and fiducial markers. Conversely, the DRR images also contained anatomical details that were absent from the FPD images, contributing to the observed discrepancies. We evaluated image quality using three standard metrics: MAE, PSNR, and SSIM. In contrast, Karbhari *et al.* (35) used Inception Score (IS) and Fréchet Inception Distance (FID) (36) to assess the quality of generated images with a deep neural network (DNN) (37). However, IS and FID require a large number of test cases to provide statistically reliable results. Due to the limited number of test cases available in our study, we chose not to use these metrics for image quality evaluation.

Recent publications have shown that incorporating a self-attention mechanism into U-Net can improve performance compared to the standard U-Net model (38-40). Although we did not apply self-attention in our study, its implementation could potentially enhance the quality of synthetic FPD images generated by our DNN.

## IV.B. CycleGAN: Prior Work and Our Style Transfer Framework

Several studies on medical imaging applications of CycleGAN, similar to those used in our work, have been reported. Lei et al. developed a CycleGAN-based network to generate CT images from MRI data (41). In their study, paired MRI and CT scans were used, and the originally unsupervised CycleGAN method was enhanced by incorporating a distance loss function that combines LP-norm distance and gradient difference. Tien et al. proposed the Cycle-Deblur GAN, which integrates CycleGAN with Deblur-GAN to improve the image quality of chest CBCT scans (42). By using paired CBCT and MSCT images and introducing additional constraints—such as MAE and Sobel filter loss—between generated and ground-truth images, the Cycle-Deblur GAN effectively mitigated the impact of misalignment errors caused by differences in acquisition dates, achieving high-quality chest image generation beyond the capabilities of CycleGAN alone.

In image-to-image translation (I2I) with GANs, there is often concern that essential image content may be lost during the transformation process. Consequently, as shown in previous studies, explicit constraints are frequently introduced to preserve critical features. In our research, we leverage the fact that DRR images and FPD images serve as paired data for treatment comparisons, and we integrate cycle consistency within a supervised learning framework. Specifically, we introduce a feature-based style loss (43) between the synthetic FPD images (produced from DRR images) and the ground-truth FPD images. Our objective is to transform DRR images into FPD-quality images while preserving the underlying structure of the DRR images, effectively treating the task as a form of style transfer. To achieve this, we employ a pre-trained VGG19 model to capture the "style" of FPD images and incorporate the style loss between the synthetic and ground-truth FPD images into our CycleGAN approach.

### IV.C. The potential application of our DNN

One challenge in patient setup verification is that the 2D–3D image registration software used to align FPD images with reference DRR images still requires considerable time. While the calculation itself takes only about 10–20 seconds, verifying the position can take 2–5 minutes. Moreover, differences in the image quality of various modalities may make visual comparison difficult. One potential solution is to convert both images into the same modality by applying our DNN.

We have already treated the thoracoabdominal region using markerless tracking with the multiplate matching method (19). To prepare template images for this approach, FPD images must be acquired over a few respiratory cycles before treatment. Because these FPD images do not include the tumor position, medical staff must manually input the tumor location—a time-consuming process. On the other hand, 4D-DRR images already contain the tumor position. By adjusting the quality of 4D-DRR images to match that of FPD images, synthetic FPD images that include the tumor position can be generated and used as template images. As a result, the preparation time is reduced, thereby improving overall treatment throughput.

Another potential application of our DNN is to expand chest X-ray datasets for deep learning by increasing the amount of training data. By converting DRR images into FPD-quality images, this approach enables efficient generation of high-quality datasets while excluding unnecessary structures, such as irradiation ports. This can help improve data consistency and enhance both the accuracy and reliability of deep learning models.

## IV.D. Study limitation

This study has a few limitations. First, the training and test data used in this study were both acquired at our hospital using the same X-ray imaging systems. It is well known that a model trained on a specific dataset may perform poorly when applied to data from a different dataset acquired with a different imaging system due to variations in image quality, such as contrast and noise, a phenomenon referred to as domain shift. To ensure the broader applicability of our DNN to other hospitals, it is essential to test its performance using FPD images acquired from hospitals utilizing X-ray imaging system different from those used at our hospital.

Second, the current method remains inadequate for accurately visualizing implanted fiducial markers. In Figure 3c, for example, overlapping organs appear as a small dark object, which our DNN incorrectly identifies as a fiducial marker (green arrow). This error likely arises from the limited training data available for fiducial markers. While our hospital uses small, ball-shaped markers, various complex-shaped markers (e.g. GoldAncher (Gold Anchor, Naslund Medical, Stockholm, Sweden), and VisiCoil (Core Oncology, Santa Barbara, Calif)) have also become commercially available. Because not all patients are implanted with fiducial markers, gathering sufficient data to encompass these diverse marker types will take considerable time. Alternatively, further modifications to our DNN could help improve its ability to accurately render fiducial markers.

## CONCLUSION

In this study, we developed a DNN to generate FPD images from DRR images of the lung region and evaluated their image quality. Despite some positional misalignment in the training data, the synthetic FPD images closely resembled the actual FPD images, and the generation process was sufficiently fast for practical use. By converting DRR images into FPD images, our approach has the potential to streamline image-guided radiotherapy and enhance clinical workflow. Moving forward, testing this method on images acquired under varied conditions at multiple centers will be essential for establishing a robust, generalizable model and facilitating broader clinical application.

**REFERENCES**

1.      Vincent P, Larochelle H, Lajoie I, Bengio Y, Manzagol PA. Stacked Denoising Autoencoders: Learning Useful Representations in a Deep Network with a Local Denoising Criterion. Journal of Machine Learning Research. 2010;11:3371-408.

2.      Jonathan M, Ueli M, Dan C, Jürgen S. Stacked convolutional auto-encoders for hierarchical feature extraction.    International Conference on Artificial Neural Networks; Heidelberg: Springer Berlin; 2011. p. 52-9.

3.      Yang W, Chen Y, Liu Y, Zhong L, Qin G, Lu Z, et al. Cascade of multi-scale convolutional neural networks for bone suppression of chest radiographs in gradient domain. Med Image Anal. 2017;35:421-33.

4.      Dong X, Lei Y, Wang TH, Thomas M, Tang L, Curran WJ, et al. Automatic multiorgan segmentation in thorax CT images using U-net-GAN. Medical Physics. 2019;46(5):2157-68.

5.      Wang J, Lu J, Qin G, Shen L, Sun Y, Ying H, et al. Technical Note: A deep learning-based autosegmentation of rectal tumors in MR images. Med Phys. 2018;45(6):2560-4.

6.      Shen CY, Nguyen D, Chen LY, Gonzalez Y, McBeth R, Qin N, et al. Operating a treatment planning system using a deep-reinforcement learning-based virtual treatment planner for prostate cancer intensity-modulated radiation therapy treatment planning. Medical Physics. 2020;47(6):2329-36.

7.      Liang X, Chen L, Nguyen D, Zhou Z, Gu X, Yang M, et al. Generating synthesized computed tomography (CT) from cone-beam computed tomography (CBCT) using CycleGAN for adaptive radiation therapy. Phys Med Biol. 2019;64(12):125002.

8.      de Vos BD, Berendsen FF, Viergever MA, Sokooti H, Staring M, Isgum I. A deep learning framework for unsupervised affine and deformable image registration. Med Image Anal. 2019;52:128-43.

9.      Hirai R, Sakata Y, Tanizawa A, Mori S. Real-time tumor tracking using fluoroscopic imaging with deep neural network analysis. Phys Medica. 2019;59:22-9.

10.     Takahashi W, Oshikawa S, Mori S. Real-time markerless tumour tracking with patient-specific deep learning using a personalised data generation strategy: proof of concept by phantom study. Br J Radiol. 2020;93(1109):20190420.

11.     Xiang L, Wang Q, Nie D, Zhang L, Jin X, Qiao Y, et al. Deep embedding convolutional neural network for synthesizing CT image from T1-Weighted MR image. Med Image Anal. 2018;47:31-44.

12.     Zhao B, Cheng T, Zhang X, Wang J, Zhu H, Zhao R, et al. CT synthesis from MR in the pelvic area using Residual Transformer Conditional GAN. Comput Med Imaging Graph. 2023;103:102150.

13.     Armanious K, Jiang C, Fischer M, Kustner T, Hepp T, Nikolaou K, et al. MedGAN: Medical image translation using GANs. Comput Med Imaging Graph. 2020;79:101684.

14.     Dong X, Wang T, Lei Y, Higgins K, Liu T, Curran WJ, et al. Synthetic CT generation from

non-attenuation corrected PET images for whole-body PET imaging. Phys Med Biol. 2019;64(21):215016.

15.    Lei Y, Dong X, Tian Z, Liu Y, Tian S, Wang T, et al. CT prostate segmentation based on synthetic MRI-aided deep attention fully convolution network. Med Phys. 2020;47(2):530-40.

16.    Abu-Srhan A, Almallahi I, Abushariah MAM, Mahafza W, Al-Kadi OS. Paired-unpaired Unsupervised Attention Guided GAN with transfer learning for bidirectional brain MR-CT synthesis. Comput Biol Med. 2021;136:104763.

17.    Mori S, Hirai R, Sakata Y, Tachibana Y, Koto M, Ishikawa H. Deep neural network-based synthetic image digital fluoroscopy using digitally reconstructed tomography. Phys Eng Sci Med. 2023;46(3):1227-37.

18.    Mori S, Takei Y, Shirai T, Hara Y, Furukawa T, Inaniwa T, et al. Scanned carbon-ion beam therapy throughput over the first 7 years at National Institute of Radiological Sciences. Phys Medica. 2018;52:18-26.

19.    Mori S, Karube M, Shirai T, Tajiri M, Takekoshi T, Miki K, et al. Carbon-Ion Pencil Beam Scanning Treatment With Gated Markerless Tumor Tracking: An Analysis of Positional Accuracy. Int J Radiat Oncol Biol Phys. 2016;95(1):258-66.

20.    Cui Y, Dy JG, Sharp GC, Alexander B, Jiang SB. Multiple template-based fluoroscopic tracking of lung tumor mass without implanted fiducial markers. Phys Med Biol. 2007;52(20):6229-42.

21.    Zhu JY, Park T, Isola P, Efros AA, editors. Unpaired Image-to-Image Translation Using Cycle-Consistent Adversarial Networks. 2017 IEEE International Conference on Computer Vision (ICCV); 2017 22-29 Oct. 2017.

22.    Mori S, Shirai T, Takei Y, Furukawa T, Inaniwa T, Matsuzaki Y, et al. Patient handling system for carbon ion beam scanning therapy. J Appl Clin Med Phys. 2012;13(6):3926.

23.    Mori S, Kumagai M, Miki K, Fukuhara R, Haneishi H. Development of fast patient position verification software using 2D-3D image registration and its clinical experience. J Radiat Res. 2015;56(5):818-29.

24.    Mori S, Shibayama K, Tanimoto K, Kumagai M, Matsuzaki Y, Furukawa T, et al. First clinical experience in carbon ion scanning beam therapy: retrospective analysis of patient positional accuracy. J Radiat Res. 2012;53(5):760-8.

25.    Ulyanov D, Vedaldi A, Lempitsky VS. Instance Normalization: The Missing Ingredient for Fast Stylization. ArXiv. 2016;abs/1607.08022.

26.    Radford A, Metz L, Chintala S. Unsupervised Representation Learning with Deep Convolutional Generative Adversarial Networks. arXiv:150806576. 2015.

27.    Mori S, Inaniwa T, Kumagai M, Kuwae T, Matsuzaki Y, Furukawa T, et al. Development of digital reconstructed radiography software at new treatment facility for carbon-ion beam scanning of National Institute of Radiological Sciences. Australas Phys Eng Sci Med. 2012;35(2):221-9.

28.    Vasiliadis G, Antonatos S, Polychronakis M, Markatos E, Ioannidis S. Gnort: High

Performance Network Intrusion Detection Using Graphics Processors. Proceedings of the 11th International Symposium on Recent Advances in Intrusion Detection (RAID). 2008:116-34.

29.　Kingma D, Ba J. Adam: A Method for Stochastic Optimization. In Proceedings of the 3rd International Conference on Learning Representations (ICLR)2015.

30.　Simonya K, Zisserman A. Very deep convolutional networks for large-scale image recognition. International Conference for Learning Representations2015.

31.　Wang Z, Bovik AC, Sheikh HR, Simoncelli EP. Image quality assessment: from error visibility to structural similarity. IEEE Trans Image Process. 2004;13(4):600-12.

32.　Bińkowski M, Sutherland DJ, Arbel M, Gretton A. Demystifying MMD GANs2018 January 01, 2018:[arXiv:1801.01401 p.]. Available from: https://ui.adsabs.harvard.edu/abs/2018arXiv180101401B.

33.　Harms J, Lei Y, Wang T, Zhang R, Zhou J, Tang X, et al. Paired cycle-GAN-based image correction for quantitative cone-beam computed tomography. Med Phys. 2019;46(9):3998-4009.

34.　Dai X, Lei Y, Fu Y, Curran WJ, Liu T, Mao H, et al. Multimodal MRI synthesis using unified generative adversarial networks. Med Phys. 2020;47(12):6343-54.

35.　Salimans T, Goodfellow I, Zaremba W, Cheung V, Radford A, Chen X. Improved techniques for training gans. Advances in neural information processing systems. 2016;29.

36.　Heusel M, Ramsauer H, Unterthiner T, Nessler B, Hochreiter S. Gans trained by a two time-scale update rule converge to a local nash equilibrium. Advances in neural information processing systems. 2017;30.

37.　Karbhari Y, Basu A, Geem ZW, Han GT, Sarkar R. Generation of Synthetic Chest X-ray Images and Detection of COVID-19: A Deep Learning Based Approach. Diagnostics (Basel). 2021;11(5).

38.　Cao H, Wang Y, Chen J, Jiang D, Zhang X, Tian Q, et al., editors. Swin-unet: Unet-like pure transformer for medical image segmentation. European conference on computer vision; 2022: Springer.

39.　Schlemper J, Oktay O, Schaap M, Heinrich M, Kainz B, Glocker B, et al. Attention gated networks: Learning to leverage salient regions in medical images. Med Image Anal. 2019;53:197-207.

40.　Saha S, Dutta S, Goswami B, Nandi D. ADU-Net: An Attention Dense U-Net based deep supervised DNN for automated lesion segmentation of COVID-19 from chest CT images. Biomed Signal Process Control. 2023;85:104974.

41.　Lei Y, Harms J, Wang T, Liu Y, Shu HK, Jani AB, et al. MRI-only based synthetic CT generation using dense cycle consistent generative adversarial networks. Med Phys. 2019;46(8):3565-81.

42.　Tien HJ, Yang HC, Shueng PW, Chen JC. Cone-beam CT image quality improvement using Cycle-Deblur consistent adversarial networks (Cycle-Deblur GAN) for chest CT imaging in breast cancer patients. Sci Rep. 2021;11(1):1133.

43.    Gatys LA, Ecker AS, Bethge M, editors. Image style transfer using convolutional neural networks. Proceedings of the IEEE Conference on Computer Vision and Pattern Recognition (CVPR); 2016.

**Figure legend**

**Table 1**

Image quality assessment with the DNN averaged over all patients.

(Mean ± SD)

|  | MAE | PSNR [dB] | SSIM | KID | Time [msec] |
|---|---|---|---|---|---|
| DRR | 0.32 ± 0.09 | 9.60 ± 2.52 | 0.32 ± 0.16 | $5.1 \times 10^{-2}$ | NA |
| U-Net | 0.24 ± 0.08 | 12.62 ± 3.26 | 0.35 ± 0.16 | $1.7 \times 10^{-2}$ | 139.6 ± 5.5 |
| Ours | 0.06 ± 0.03 | 24.47 ± 5.09 | 0.69 ± 0.16 | $0.7 \times 10^{-2}$ | 12.9 ± 5.2 |

*Abbreviations: SD = standard deviation; DRR = digitally reconstructed radiography; MAE mean absolute error; PSNR = peak signal-to-noise ratio; SSIM = structural similarity index measure; KID = Kernel Inception Distance.*

**Figure 1**

The proposed DNN consists of an image generator network that generates FPD images from DRR images (G<sub>DRR-FPD</sub>), an image generator network that generates DRR images from FPD images (G<sub>FPD-DRR</sub>), as well as a Discriminator and the VGG19 network. The DRR image is input into both G<sub>DRR-FPD</sub> and G<sub>FPD-DRR</sub>, producing a synthetic FPD image and an identical DRR image, respectively. The synthetic FPD image is further input into (G<sub>FPD-DRR</sub>) to generate a cycle DRR image. The cycle-consistency loss is calculated between the input DRR and the cycle DRR images. The identity loss is computed using the input DRR and the identical DRR images. Adversarial loss is determined by comparing the synthetic FPD with the ground-truth FPD images. Style loss is calculated between the synthetic FPD and the ground-truth FPD images.

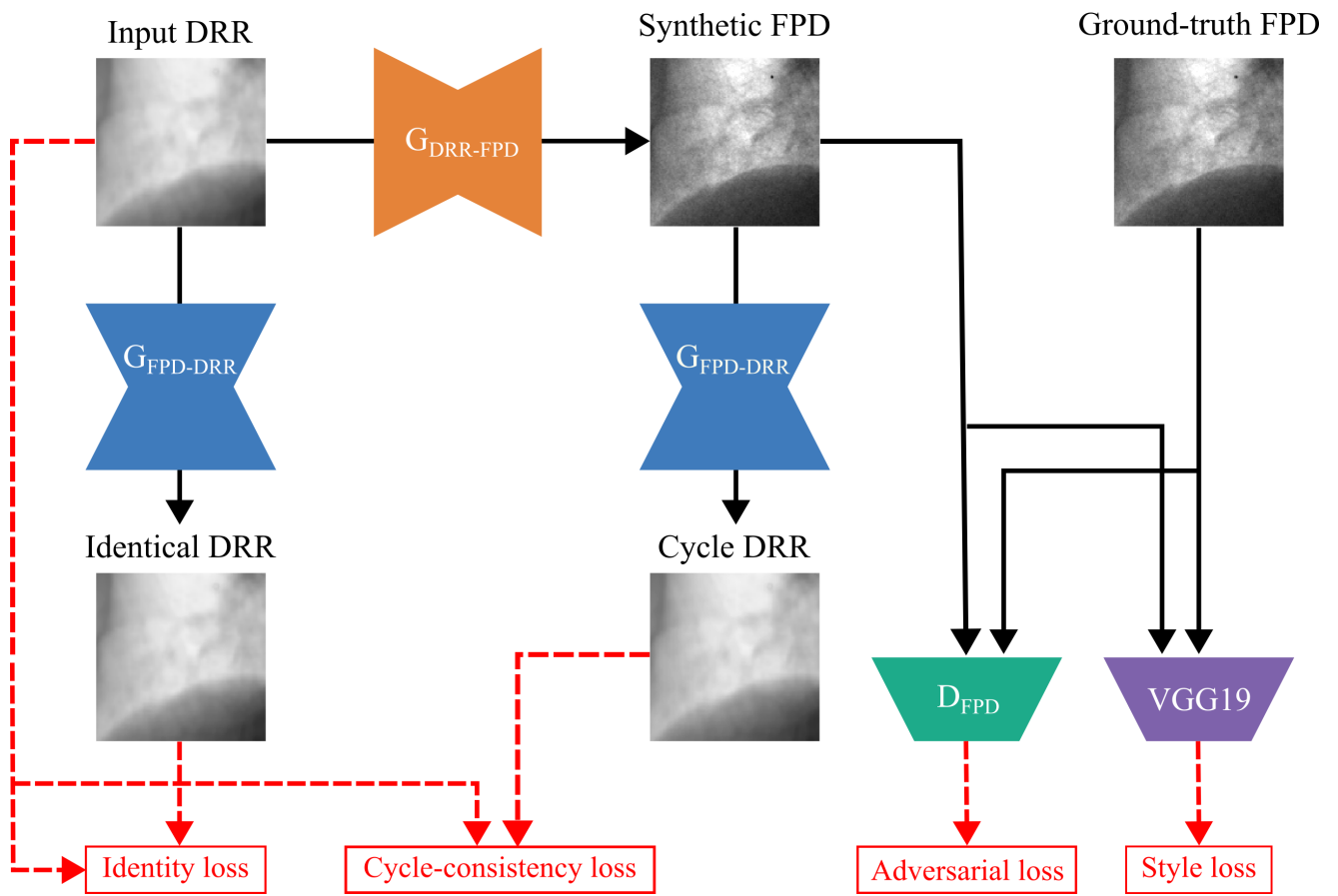

*Abbreviations: DRR = digitally reconstructed radiography; DNN = deep neural network; FPD = flat panel detector.*

**Figure 2**

Network structures for (a) the generator deep neural network and (b) the discriminator. Convolutional kernel size, stride size and the number of output channels are expressed by (kernel, stride, outputs channels) in the figures.

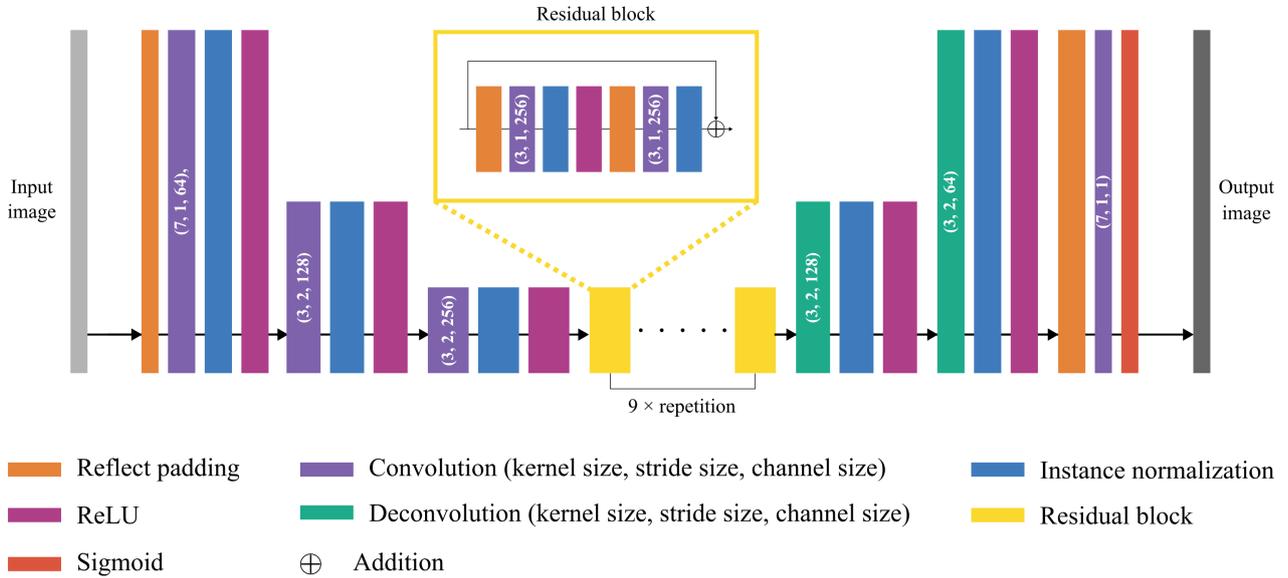

(a)

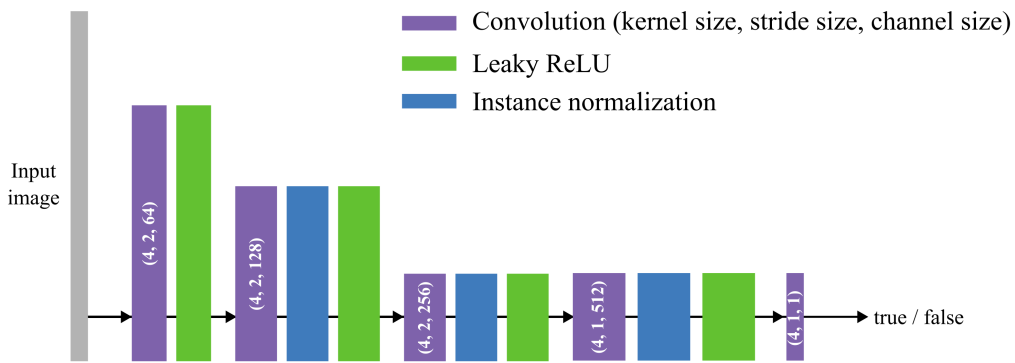

(b)

*Abbreviations: ReLU = Rectified Linear Unit*

**Figure 3**

Resultant images for case no.4. (a) Input DRR image. (b) Ground-truth FPD image. Blue and maganda arrows show the irradiation port cover edge and the patient call cable, respectively. (c) Synthetic FPD image with our DNN. (d) Synthetic FPD image with the U-Net. Yellow and green arrows show the implanted fiducial markers.

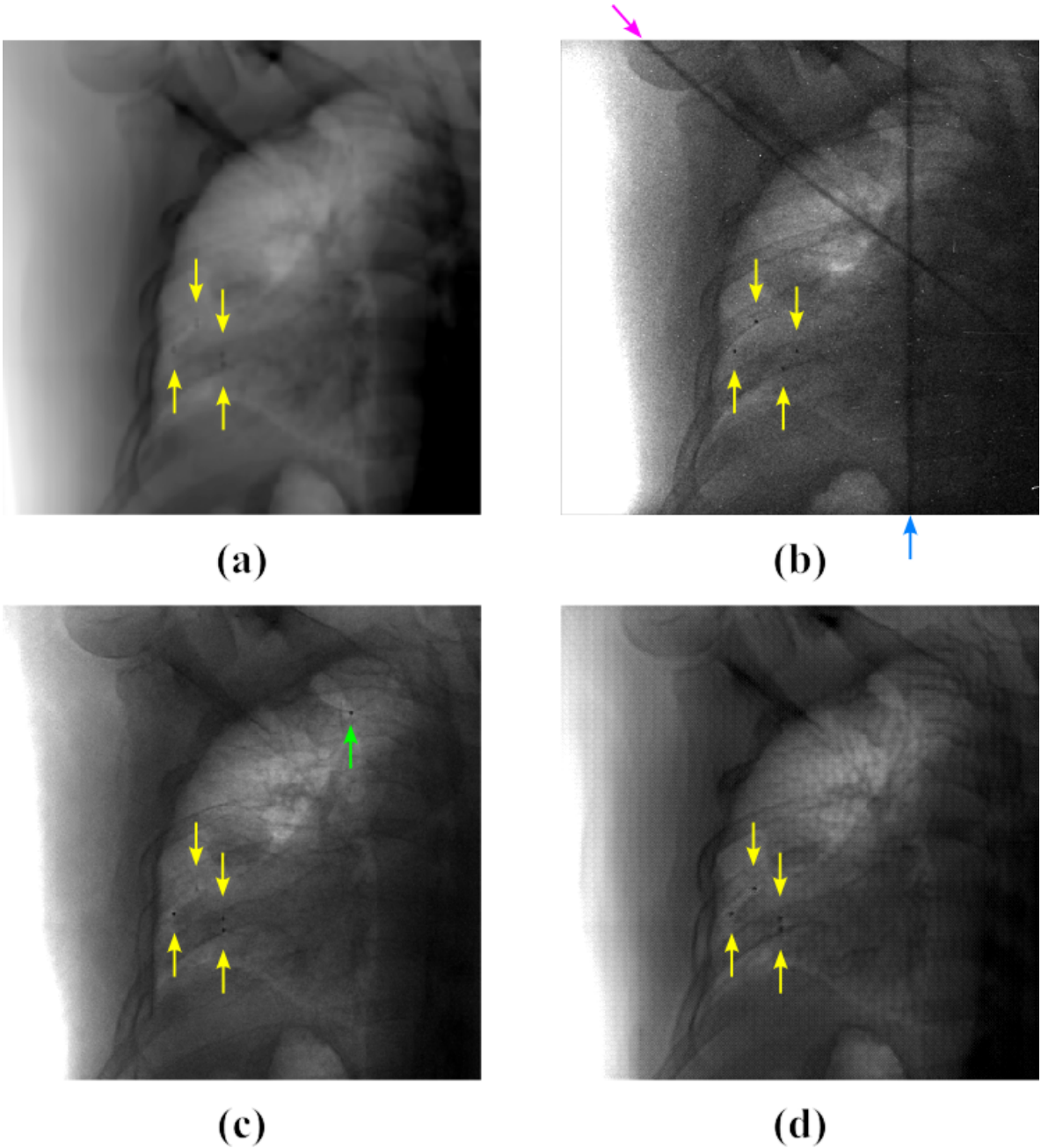

**Figure 4**

Resultant images for case no.6. (a) Input DRR image. (b) Ground-truth FPD image. Maganda arrow shows the patient call cable. (c) Synthetic FPD image with our DNN. (d) Synthetic FPD image with the U-Net.

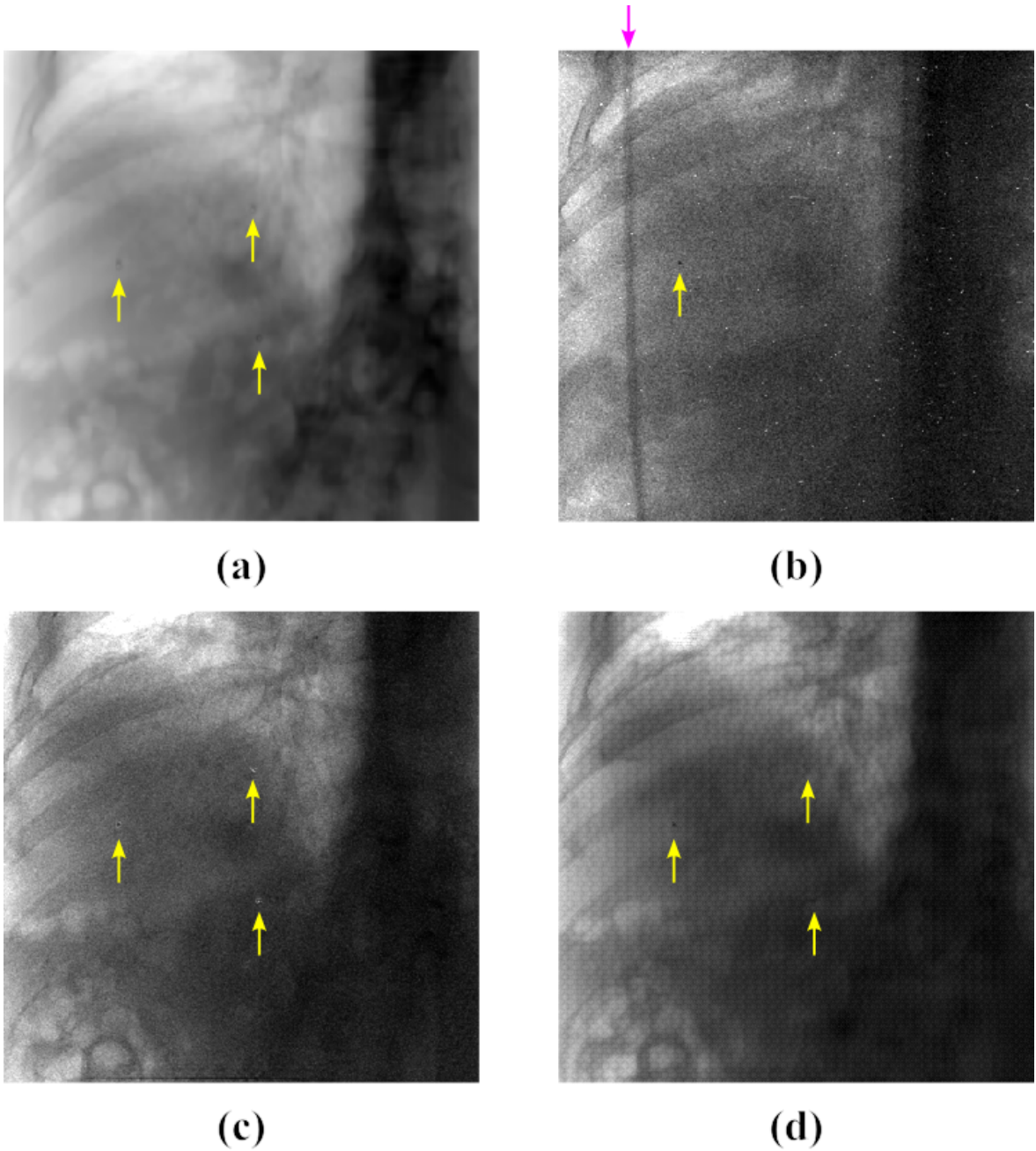

**Figure 5**

Image quality metrics for the synthetic FPD and DRR with the ground-truth FPD images. (a) MSE, (b) PSNR and (c) SSIM. The horizontal line in the center of the box represents the median, while the bottom and top edges of the box correspond to the 25th percentile ($q1$) and 75th percentile ($q3$), respectively. The whiskers extend to the furthest data points within the non-outlier range. Outliers, depicted as light blue open circles, are defined as values greater than $q3 + 1.5 \times (q3 - q1)$ or less than $q1 - 1.5 \times (q3 - q1)$.

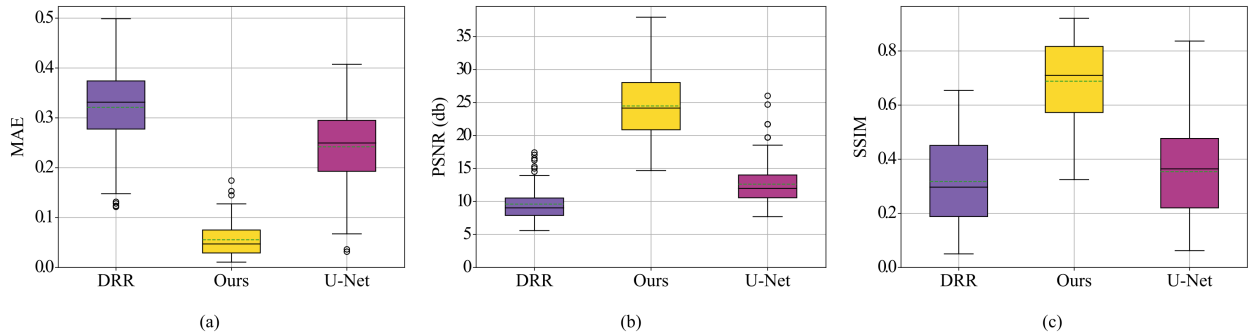

*Abbreviations: DRR = digitally reconstructed radiography; FPD = flat panel detector; MSE mean square error; PSNR = peak signal-to-noise ratio; SSIM = structural similarity index measure.*